\documentclass[
]{ceurart}

\usepackage{listings}
\usepackage{graphicx}
\usepackage{subcaption}
\usepackage{float}
\begin{document}

\copyrightyear{2025}
\copyrightclause{Copyright for this paper by its authors.
  Use permitted under Creative Commons License Attribution 4.0
  International (CC BY 4.0).}

\conference{LLAIS 2025: Workshop on LLM-Based Agents for Intelligent Systems, at ECAI 2025}

\title{Analyzing the Correlation Between Hallucinations and Knowledge Conflicts in Large Language Models}


\author[]{Lucrezia Laraspata}[%
orcid=0009-0003-8136-9140,
email=l.laraspata3@phd.uniba.it,
]
\cormark[1]

\address[]{Department of Computer Science, University of Bari Aldo Moro, Bari, Italy}

\author[]{Giovanna Castellano}[%
orcid=0000-0002-6489-8628,
email=giovanna.castellano@uniba.it,
]

\author[]{Gennaro Vessio}[%
orcid=0000-0002-0883-2691,
email=gennaro.vessio@uniba.it,
]


\cortext[1]{Corresponding author.}

\begin{abstract}
Hallucinations, factually incorrect or unverifiable answers, represent one of the most challenging limitations of Large Language Models (LLMs), particularly for knowledge-intensive tasks. According to several studies, one possible explanation lies in the internal \textit{knowledge conflicts} that arise due to the models’ fixed and outdated training data. In this paper, we investigate whether internal representations associated with knowledge conflicts are correlated with hallucination behaviors in LLMs. Using \textit{probing techniques} inspired by two prior works, we analyzed activations from hidden, attention, and MLP layers, as well as output logits, across predefined tasks to assess the correlation between the two phenomena. Specifically, we probe LLaMA-3-8B on hallucination detection benchmarks and Falcon-7B on a dataset of knowledge conflicts. Our findings indicate that, although these phenomena appear conceptually related, the internal activation patterns of hallucinations cannot be fully reduced to, or explained by, knowledge conflict representations. Nonetheless, we observe that probing remains a robust tool across multiple languages and activation types, supporting its use for improving LLMs' interpretability. This work contributes to the broader understanding of hallucinations in LLMs and highlights the importance of fine-grained analysis of their internal behavior. The source code is publicly available at: \url{https://github.com/llaraspata/HallucinationDetection}.
\end{abstract}


\maketitle

\section{Introduction}

Despite their impressive performance across a wide range of tasks, Large Language Models (LLMs) still exhibit notable limitations, particularly in high-stakes or knowledge-intensive domains. One key issue is that LLMs possess a fixed knowledge base limited to the data available at training time. Once these models are trained, their internal knowledge is never updated. Addressing this limitation typically requires either continual retraining or the injection of external knowledge at inference time. However, when LLMs encounter information that falls outside their limited knowledge boundaries, they are more susceptible to generating hallucinations (i.e., factually incorrect, inconsistent, or nonsensical outputs \cite{sun2025hallucinations}) as they often resort to fabricating facts or providing answers that were correct in the past but are now outdated \cite{huang2025survey}. Compounding this, recent studies show that LLMs often prioritize contextual cues over factual accuracy, even in the presence of misinformation, reducing the effectiveness of fine-tuning or prompt-based mitigation strategies \cite{xie2023adaptive}.

In their survey \cite{huang2025survey}, Huang et al.~introduce a more intuitive taxonomy of hallucinations in LLMs, dividing them into two principal categories: factuality hallucinations and faithfulness hallucinations. This study focuses on \textit{factuality hallucinations}, which occur when an LLM generates outputs that are either inconsistent with real-world facts or unverifiable \cite{li2024dawn}. They can be further classified into factual contradictions and factual fabrications. The first ones occur when the generated content appears to be grounded in reality but contains internal inconsistencies. These typically result from how the model captures, stores, or expresses factual knowledge, and may be further characterized as either entity-error hallucinations or relation-error hallucinations, depending on the specific type of contradiction. In contrast, factual fabrications refer to instances in which the model produces information that cannot be verified against established knowledge. These are often categorized as either unverifiability hallucinations or overclaim hallucinations, depending on whether the fabricated information is merely unverifiable or presented with unwarranted certainty.

A straightforward yet effective technique used to analyze a model's internal representations with respect to a given phenomenon is \textit{probing} \cite{belinkov2022probing}. Zhao et al.~\cite{zhao2024analysingresidualstreamlanguage} introduced a linear probing method to detect \textit{knowledge conflicts}---which happen when LLMs' parametric knowledge conflicts with the information provided in the context---within the residual stream of intermediate LLM activations. Their approach achieved high accuracy (${\sim}85\%$) in detecting such a phenomenon without requiring architectural modifications or retraining, providing an efficient representation for internal inconsistencies. Similarly, Snyder et al.~\cite{snyder2024early} utilized output logits and internal layer activations, achieving an AUROC of \(75\%\) in detecting hallucinations.

However, while these techniques effectively leverage knowledge conflicts and hallucination detection within LLMs' internal representations, the connection between the two phenomena remains underexplored. In this paper, we investigate whether internal knowledge conflict signals correlate with hallucination behaviors in factual question-answering (QA), and vice versa. Building on Zhao et al.'s linear probing framework, we analyze whether activation-level indicators of conflict align with hallucinations in generated outputs by probing residual stream activations across all hidden, attention, and MLP layers. Likewise, we employed Snyder et al.'s probing method to analyze whether signs of hallucinations, observed in both internal activations and output logits, are associated with knowledge conflicts. Our study evaluates LLaMA-3-8B \cite{grattafiori2024llama} on three benchmark datasets of hallucinated question-answer pairs, and Falcon-7B \cite{falcon40b} on a dataset of knowledge conflicts in QA, aiming to advance understanding of the factors that lead to hallucinations. We selected the LLMs that performed best in the reference research, respectively, according to their accuracy.


The rest of this paper is structured as follows. Section~\ref{sec:related_work} summarizes key contributions in hallucination detection and mitigation. Section~\ref{sec:materials} outlines the research hypothesis and design, while Section~\ref{sec:results} discusses the obtained results. Lastly, Section \ref{sec:conclusion} highlights key findings and remaining challenges.

\section{Related Work}
\label{sec:related_work}

As hallucinations present a significant challenge in generative AI, numerous research studies have been conducted to detect and mitigate them. One of the simplest strategies is the so-called ``LLM-as-a-judge'' \cite{zheng2023judging}, where an LLM is prompted to determine if there is a hallucination in the provided content. In \cite{ravi2024lynx}, the authors propose a hallucination evaluation LLM, fine-tuned to determine if the generated answer---derived from a Retrieval-Augmented Generation (RAG) pipeline---is hallucinated (or not) and explain why. Despite gathering good results, Li et al.~investigated the effectiveness of this type of strategy in \cite{li2023halueval}. They tested several LLMs (both closed and open-weight) across three different tasks (QA, dialog, and text summarization), and their results demonstrated that LLMs are still poor at identifying hallucinations.

Several studies have employed RAG techniques to mitigate hallucinations, resulting in tangible improvements in factual accuracy \cite{lewis2020retrieval, ram2023context}. However, as Barnett et al.~argue in \cite{barnett2024seven}, these methods do not fully resolve the core issue. Specifically, the generation bottleneck remains: even when grounded on accurately retrieved content, LLMs may still hallucinate due to inherent limitations in the decoding process. Furthermore, LLMs have been shown to prioritize contextual coherence over factual correctness \cite{xie2023adaptive}, which can degrade output quality when the retrieved documents contain incorrect or misleading information \cite{huang2025survey}. In such cases, retrieval may unintentionally amplify hallucinations. Finally, as highlighted by Li et al.~\cite{li2023halueval}, injecting external knowledge is often insufficient in more complex settings—such as multi-turn dialogues—where models must maintain factual consistency while navigating conversational dependencies.

Beyond these limitations, RAG-based and other retrieval-centric approaches are computationally expensive and introduce additional complexity into the inference workflow, which hinders their deployment in real-world applications. This underscores the need for lighter, more interpretable alternatives. 

Probing classifiers have recently emerged as a promising methodology for interpreting and analyzing deep neural language models \cite{belinkov2022probing}. In this framework, a lightweight classifier is trained to predict specific linguistic or cognitive properties from a model’s internal representations. Probing not only supports a finer-grained understanding of what is encoded at different layers but also contributes to reducing the opacity of learned representations \cite{conneau2018you, allen2023physics1, allen2023physics3}, making it a valuable tool for ``introspective'' analysis.

Zhao et al.~\cite{zhao2024analysingresidualstreamlanguage} proposed an analysis of how knowledge conflict signals are encoded within LLMs' activations. Specifically, they analyzed the residual stream of different activations (hidden, attention, and MLP) for all the layers of LLaMA-3-8B \cite{grattafiori2024llama}. Then, they defined a linear probing task to determine whether a knowledge conflict arose. To this end, in their paper, they presented the results obtained by training a simple logistic regressor on LLaMA-3-8B activations using the NQ-Swap dataset \cite{longpre2021nqswap}. This methodological framework enables the detection of knowledge conflict signals, which are found to emerge in intermediate layers (e.g., from the 13th to 16th layer in LLaMA-3-8B) with high accuracy.

In the study by Snyder et al.~\cite{snyder2024early}, the authors utilized a probing methodology to investigate the occurrence of hallucinations in LLMs. Specifically, they examined models such as Falcon-7B \cite{falcon40b} by probing their inputs through output logits and monitoring attention and MLP layer activations. This approach aimed to identify signs of hallucinations while addressing open-ended QA tasks using the TriviaQA dataset \cite{joshi2017triviaqa}.

The key contribution of prior works \cite{zhao2024analysingresidualstreamlanguage, snyder2024early} is that they eliminate the need for further interaction with the LLM or the construction of complex systems to retrieve external knowledge. 
However, both of them have focused either on knowledge conflicts or on hallucinations solely, without exploring their possible causes. In this work, we examine whether hallucinations and knowledge conflicts are related.

\section{Methodology and Experimental Design }
\label{sec:materials}

\subsection{Hypothesis}

This research aims to investigate whether hallucination and knowledge conflicts in representations are correlated. Formally, given an internal representation of a hallucination \(z_H\) and an internal representation of a knowledge conflict \(z_{KC}\), we want to determine if the two are comparable and encoded similarly within LLMs' internal activations.

We can reasonably assume that the LLMs were trained on accurate, real-world knowledge. Therefore, if we inject a hallucinated answer into the prompt, this creates a conflict between the hallucinated input and the LLM's internal parametric knowledge. We hypothesize that this conflict will manifest in the model's internal activations and can be used to detect hallucinations. Thus, to probe that hallucinations stem from knowledge conflicts, we employ a knowledge conflict probing model \(KC\), so that \(KC(z_H) = 1\), where \(1\) means that the pattern learned by \(KC\)  is similar to the hallucination representation \(z_H\).

Similarly, to probe that knowledge conflicts are captured as hallucinations are, we provide the model with a question and a context that conflicts with its parametric knowledge, and apply a hallucination probing model \(H\), so that \(H(z_{KC}) = 1\), where \(1\) means that the pattern learned by \(H\)  is similar to the knowledge conflict representation \(z_{KC}\).

\subsection{Datasets}

To test this hypothesis, we employed three distinct QA datasets with hallucinated answers, along with one targeting the detection of knowledge conflicts in QA.

We employed the Mu-SHROOM dataset, proposed for the homonymous challenge held at SemEval-2025 \cite{vazquez2025semeval}. The original task involved detecting spans of text corresponding to hallucinations, requiring participants to quantify how likely such spans were hallucinated. Since the dataset comprises pairs of questions and hallucinated answers, we selected a subset of relevant features from the original data: the identifier of the open-weight LLM used to generate each instance, the question serving as input, the corresponding hallucinated answer, and the language identifier used in each instance. The dataset spans 14 languages (i.e., Arabic (Modern standard), Basque, Catalan, Chinese (Mandarin), Czech, English, Farsi, Finnish, French, German, Hindi, Italian, Spanish, and Swedish), allowing us to conduct an additional analysis on model resilience across linguistic variation. As labels are provided only for the validation and test sets, our experiments were restricted to these portions, resulting in a total of \(2,401\) instances.

HaluEval \cite{li2023halueval} was the second dataset involved in our experiments. The original version contains \(30,000\) instances, divided equally into three subsets: QA, dialog, and summarization. All of them were generated by ChatGPT and then annotated by humans. For our purposes, we focused exclusively on the dialogue subset. This choice introduces greater task complexity, as it requires the models to assess hallucinations within longer conversational contexts, rather than isolated questions. 

The last hallucination dataset used is HaluBench, introduced in \cite{ravi2024lynx}, with \(14,900\) English instances sourced from real-world domains, spanning from finance to medicine. It contains tuples of (question, context, answer, label), where the label is a binary score indicating whether the answer contains a hallucination.

NQ-Swap is a synthetic dataset created by Longpre et al.~in \cite{longpre2021nqswap}, built upon the Natural Questions dataset \cite{kwiatkowski2019natural}. It consists of \(4,746\) artificially constructed conflicting data pairs designed to test and evaluate language models' ability to handle knowledge conflicts in QA tasks.

Table \ref{tab:dataset_stats} summarizes the main statistics of the datasets used in our experiments, including total instances and class distributions.

\begin{table}
    \centering
    \caption{Dataset statistics, showing the distribution of classes and the ratio of positive to negative instances. For NQ-Swap and HaluEval, each instance is annotated with answers for both labels, ensuring class balance.}
    \label{tab:dataset_stats}
    \begin{tabular}{lccc}
    \toprule
    \textbf{Dataset} & \textbf{Total Instances} & \textbf{Positive Class} & \textbf{Negative Class} \\
    \midrule
    NQ-Swap \cite{longpre2021nqswap} & \(9,492\) & \(4,746\) (Conflict) & \(4,746\) (Non-conflict)\\
    Mu-SHROOM \cite{vazquez2025semeval} & \(2,401\) & \(2,401\) (Hallucination) & \(0\) (Non-hallucination)\\
    HaluEval \cite{li2023halueval} & \(20,000\)  & \(10,000\) (Hallucination) & \(10,000\) (Non-hallucination)\\
    HaluBench \cite{ravi2024lynx} & \(14,900\) & \(7,170\) (Hallucination) & \(7,730\) (Non-hallucination)\\
    \bottomrule
    \end{tabular}
\end{table}

\subsection{Probing Framework}

The knowledge conflict probing framework proposed in \cite{zhao2024analysingresidualstreamlanguage} adopts a comprehensive methodology to investigate how LLMs manage knowledge conflicts between their parametric knowledge (stored internally) and contextual knowledge (provided in the input). Central to this approach is the detailed analysis of the residual stream within the Transformer architecture, where token embeddings are successively modified by vector additions from Self-Attention and MLP blocks across layers. 

To determine whether specific information, such as the signal of knowledge conflict or the source of knowledge reliance, is encoded within these internal activations, a linear probing technique is employed. This involves training a Logistic Regression model as a binary classifier on activations (e.g., hidden, attention, and MLP residual) to predict the presence of the probed information. The experimental setup focuses on open-domain QA tasks, utilizing datasets such as NQ-Swap, Macnoise, and ConflictQA, and models like LLaMA-3-8B and Llama2-7B. Each data instance comprises a question and conflicting or non-conflicting evidence pairs, along with corresponding answers. The model's inherent parametric knowledge is assessed in a ``closed-book'' setting, and answers are generated using a greedy decoding strategy with three in-context demonstrations for format alignment across all experiments. 

\begin{figure}[t]
    \centering
    \begin{subfigure}[b]{0.4\textwidth}
        \includegraphics[width=\textwidth]{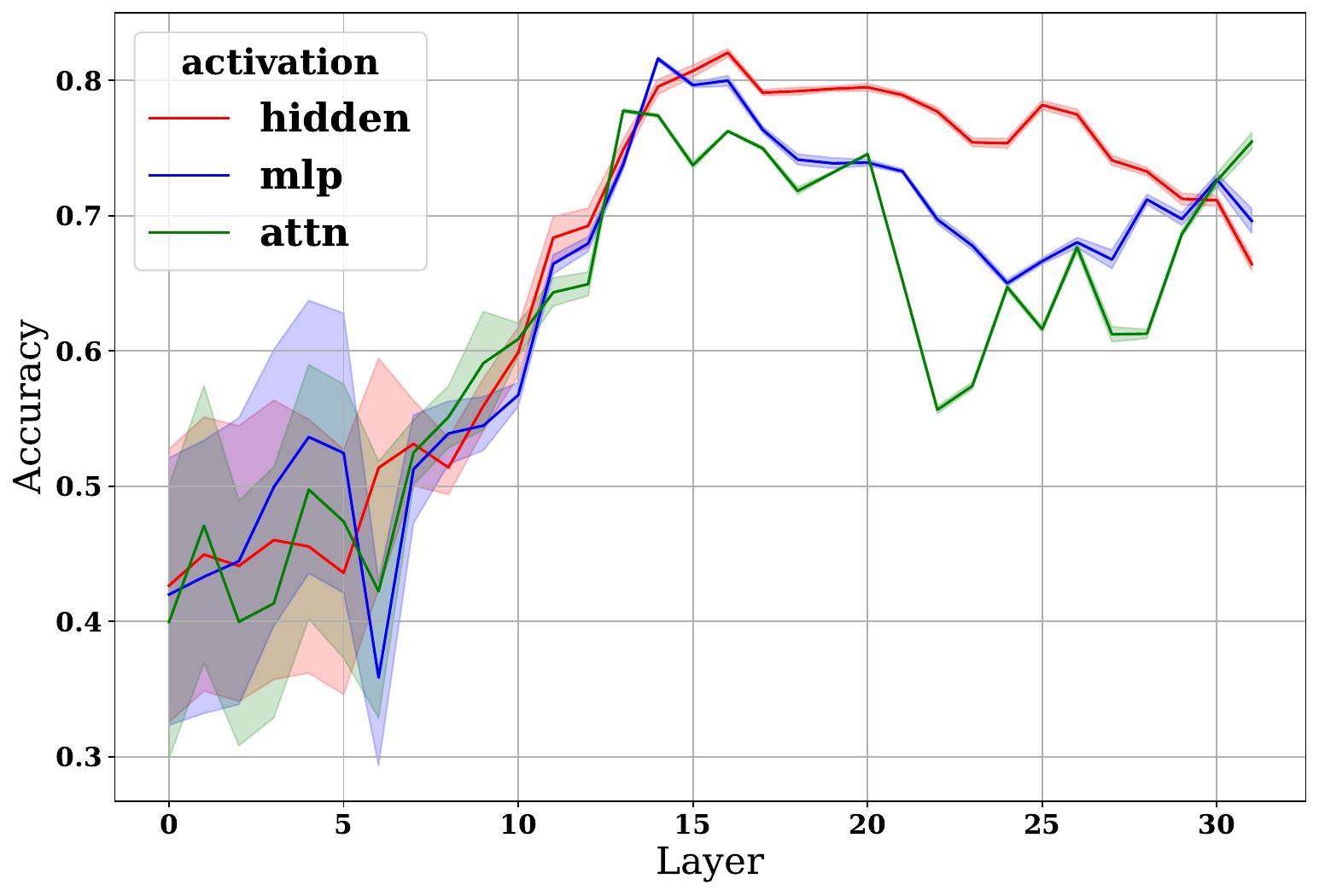} 
        \caption{Accuracy}
    \end{subfigure}
    \begin{subfigure}[b]{0.4\textwidth}
        \includegraphics[width=\textwidth]{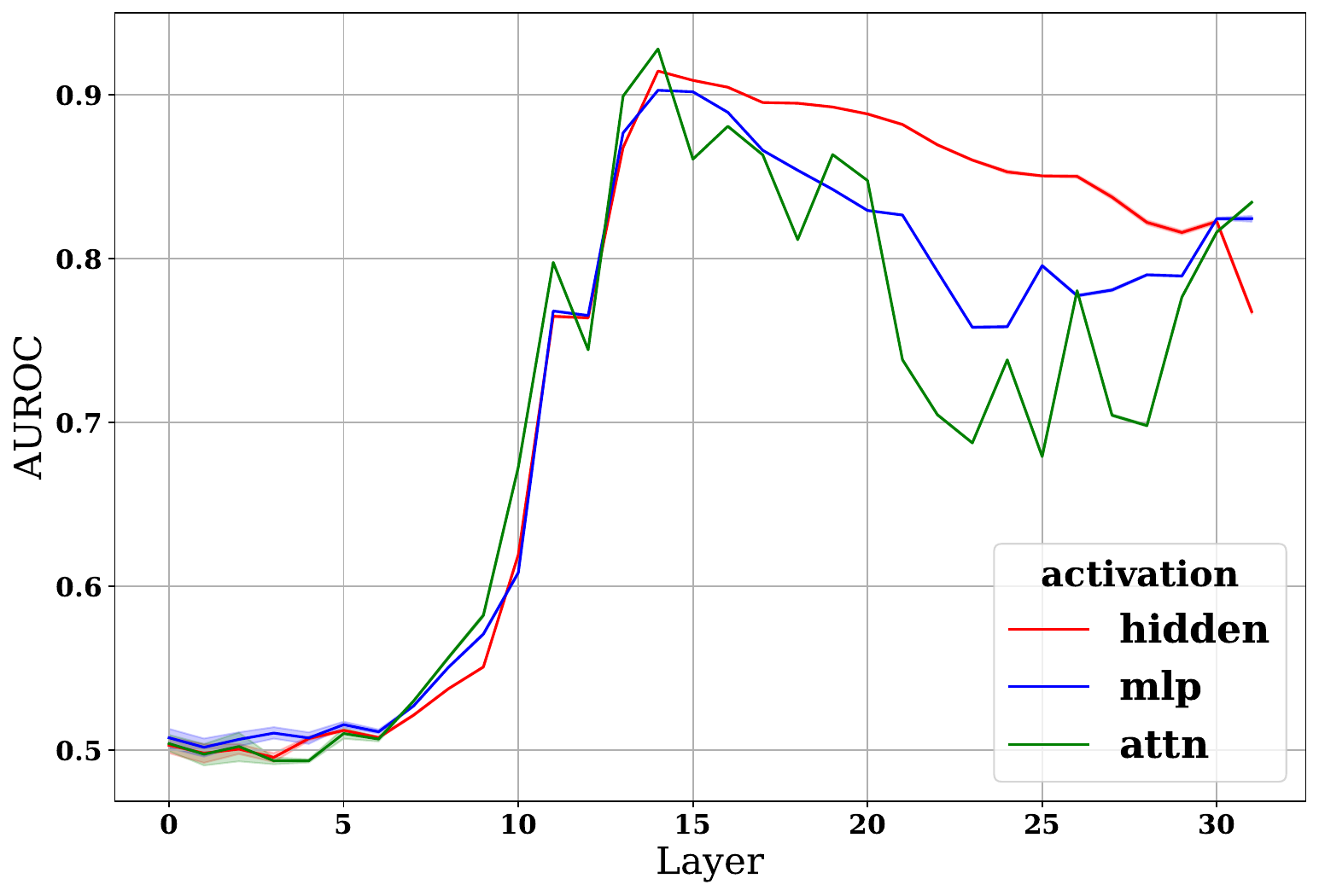}
        \caption{AUROC}
    \end{subfigure}
    \caption{
        Accuracy and AUROC of probing models on detecting knowledge conflicts based on the activations of LLaMA-3-8B. The results were obtained by reproducing the original experiments presented in \cite{zhao2024analysingresidualstreamlanguage}. Probing results for hidden, attention, and MLP activations are shown in red, green, and blue, respectively. }
    \label{fig:kc_result}
\end{figure}

As observable from Fig.~\ref{fig:kc_result}, it is possible to detect knowledge conflicts with \(\sim\)80\% accuracy at intermediate layers, namely at the 13th layer of LLaMA-3-8B. The full reproduction of their experiments, along with the training scripts and evaluation code, is available in our public repository: \url{https://github.com/llaraspata/HallucinationDetection}.

In their work \cite{snyder2024early}, Snyder et al.~propose a methodology for the detection of hallucinations in factual QA with LLMs. The core insight grounding their approach is that artifacts associated with the model's internal generation process can encode signals of hallucination. Specifically, their method involves probing LLMs at various layers to investigate three categories of these internal artifacts: attention and MLP activations extracted from all layers, and output logits. They proceed to train binary classifiers using these artifacts as input features to classify model generations as either hallucinated or non-hallucinated. A notable finding of this methodology is its capability to predict subsequent hallucinations even before they occur, specifically by analyzing artifacts associated with the first generated token. This holds even for seemingly ``uninformative'' initial tokens, such as a newline character, because the associated artifacts are high-dimensional vectors that capture the model's internal state after processing the full input. 

\begin{figure}[t]
    \centering
    \begin{subfigure}[b]{0.4\textwidth}
        \includegraphics[width=\textwidth]{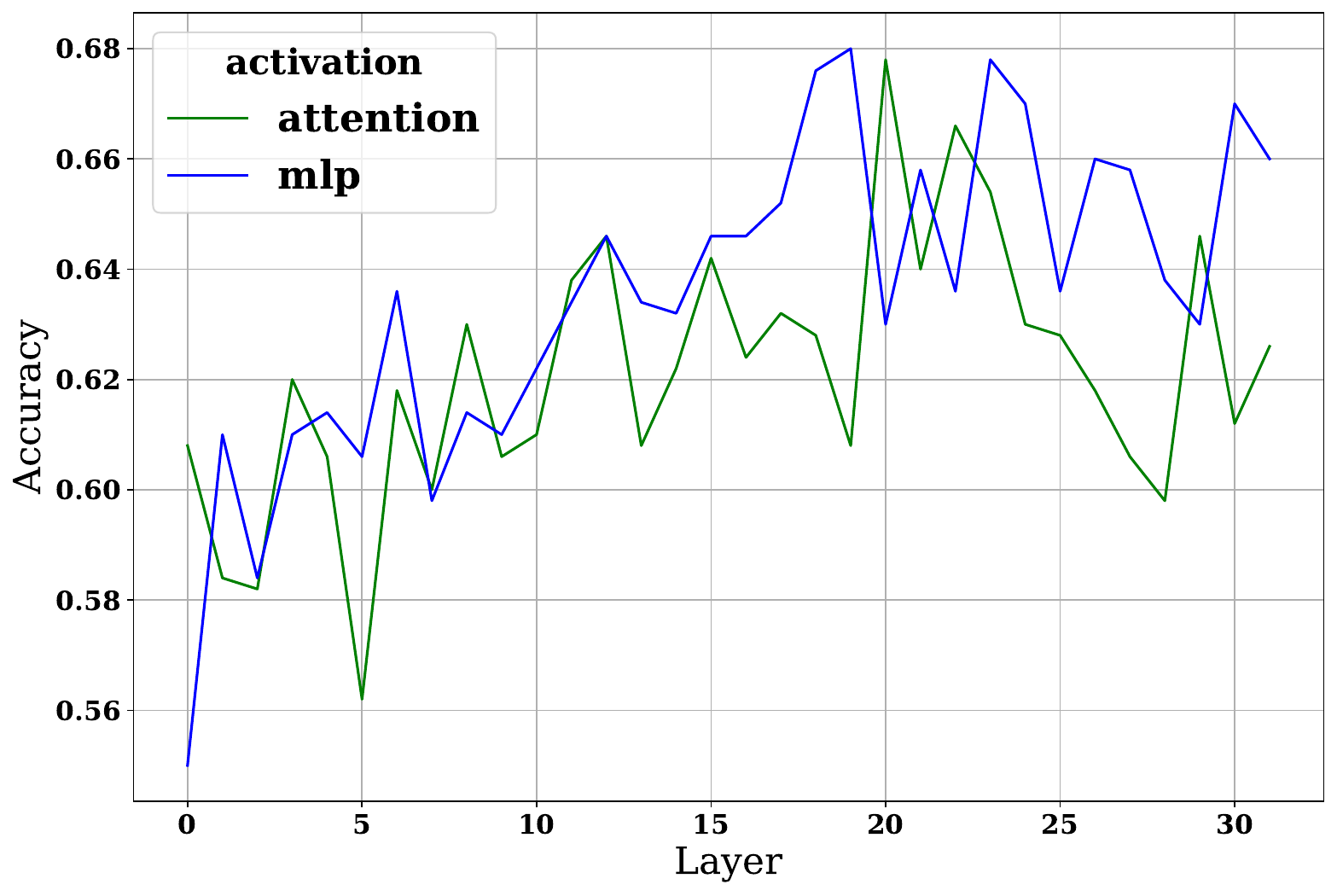} 
        \caption{Accuracy}
    \end{subfigure}
    \begin{subfigure}[b]{0.4\textwidth}
        \includegraphics[width=\textwidth]{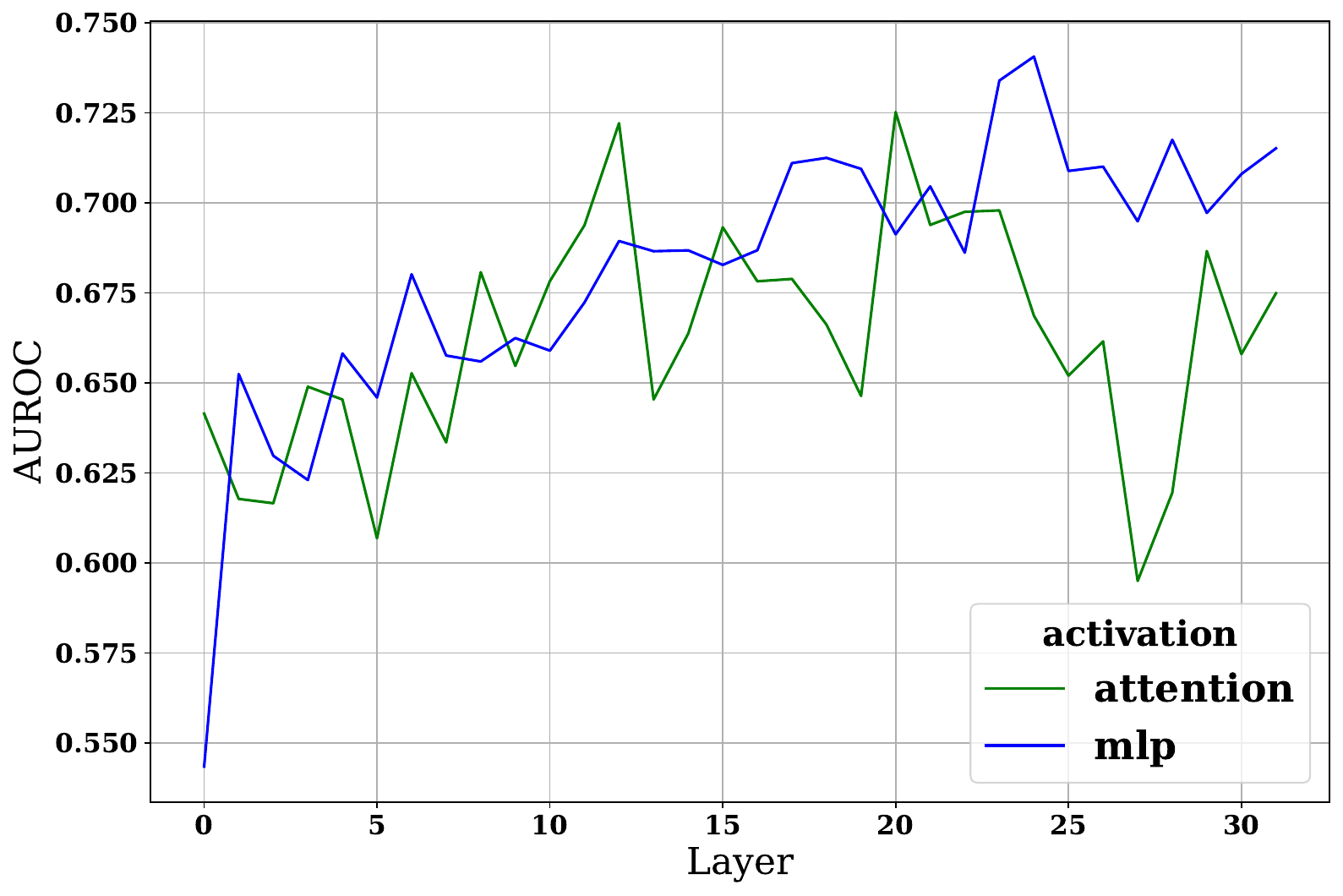}
        \caption{AUROC}
    \end{subfigure}
    \caption{
        Accuracy and AUROC of probing models on detecting hallucinations based on the activations of Falcon-7B. The results were obtained by reproducing the original experiments presented in \cite{snyder2024early}. Probing results for attention and MLP activations are shown in green and blue, respectively. }
    \label{fig:hallu_result}
\end{figure}

They showed that the distributions of these artifacts tend to differ between hallucinated and non-hallucinated generations, achieving up to \(68\%\) of accuracy, as shown in Fig.~\ref{fig:hallu_result}. This approach offers a lightweight detection mechanism that does not necessitate fine-tuning the entire LLM or repeated sampling, differentiating it from some alternative methods. The full reproduction of their experiments, along with the training scripts and evaluation code, is available in our public repository: \url{https://github.com/llaraspata/HallucinationDetection}.

\subsection{Experimental Setup}

To validate our hypothesis, we designed two experimental setups.

\subsubsection{From Knowledge Conflicts to Hallucinations}

To probe whether knowledge conflicts can predict hallucinations, we first reproduced the experiments from \cite{zhao2024analysingresidualstreamlanguage}, training knowledge conflict probing models \(KC_{a,i}\), where \(a\) denotes the activation type (hidden, attention, or MLP), and \(i\) the layer index in the range \([1, 32]\). We adopted Logistic Regression as the classifier and used the NQ-Swap dataset for training. Activations were extracted from LLaMA-3-8B \cite{grattafiori2024llama}, following the same protocol as in the original study.

To evaluate the hypothesis, we employed three hallucination-related datasets: Mu-SHROOM, HaluEval, and HaluBench. Each dataset provides tuples \((q, a, l)\), where \(q\) is a factual question, \(a\) is the answer generated by an LLM, and \(l\) is a binary label indicating hallucination (\(1\) if hallucinated, \(0\) otherwise). Note that for Mu-SHROOM, all answers are labeled as hallucinated (\(l=1\)).

We adopted a simple yet effective prompting strategy: the LLM is presented with a question \(q\) and an answer \(a\) (produced by a different model). Regardless of its output, for each layer \(i\), we extracted activations \(h_{a,i}\) corresponding to the hidden, attention, and MLP components. These activations were then passed to the corresponding \(KC_{a,i}\) model trained on that layer and activation type. Finally, we computed the accuracy between \(KC(h_{a,i})\) and the ground-truth hallucination label \(l\), to assess the predictive power of knowledge conflicts.

\begin{figure}[t]
    \centering
    \includegraphics[width=\linewidth]{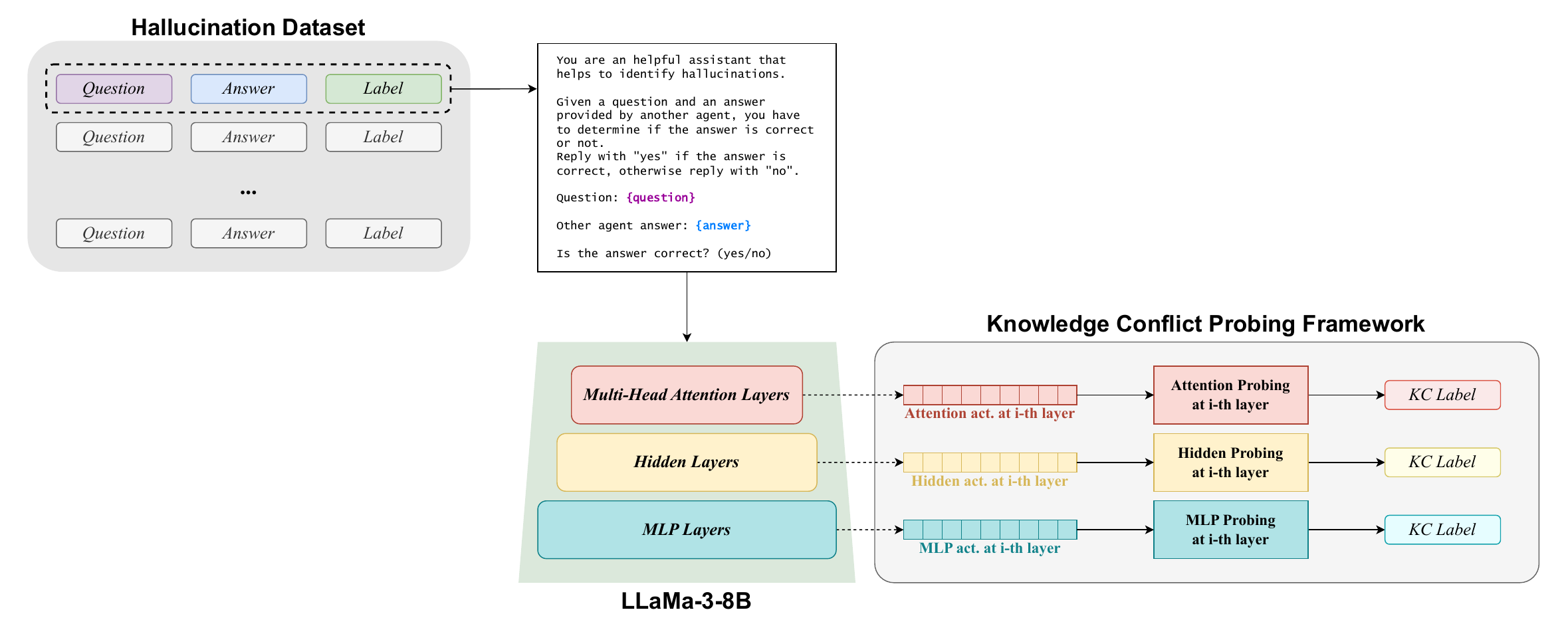}
    \caption{Probing hallucinations through knowledge conflicts. Each hallucination-related dataset is processed using LLaMA-3-8B, extracting hidden, attention, and MLP activations. These are fed into dedicated probing classifiers trained to predict knowledge conflict (KC) labels, which are then compared against the dataset-provided hallucination labels.}
    \label{fig:hall_by_kc}
\end{figure}

The overall framework is illustrated in Fig.~\ref{fig:hall_by_kc}.

\subsubsection{From Hallucinations to Knowledge Conflicts}

To assess whether hallucinations can predict knowledge conflicts, we reproduced the setup from \cite{snyder2024early}, training hallucination probing models on two types of signals: output logits \(H_l\), and intermediate activations \(H_{a,i}\), where \(a\) is either attention or MLP and \(i \in [1, 32]\). The models were trained on TriviaQA \cite{joshi2017triviaqa}, using Falcon-7B \cite{falcon40b} as the LLM and a feed-forward network as the probing classifier.

For evaluation, we used the NQ-Swap dataset, which contains tuples of the form \((q, c_p, c_c, a_p, a_c)\), where \(q\) is a factual question, \(c_p\) is a context that supports the LLM’s parametric knowledge, \(c_c\) is a context that conflicts with it, \(a_p\) is the answer aligned with the supporting context, and \(a_c\) is the answer aligned with the conflicting context. Each tuple enables the construction of two distinct prompts: one using the supporting context \(c_p\), which is assigned the label \(l = 0\), and one using the conflicting context \(c_c\), labeled as \(l = 1\). The goal is to assess whether the internal activations of the model encode the presence of a knowledge conflict. For each prompt, we recorded the model’s output logits, denoted as \(h_l\), along with the intermediate activations \(h_{a,i}\), extracted from the attention and MLP components across all layers. These features were then passed to the probing models \(H_l\) and \(H_{a,i}\), respectively. Accuracy was computed by comparing the predicted conflict label with the actual label \(l\), thus quantifying the strength of the implication from hallucination to knowledge conflict.

\begin{figure}[t]
    \centering
    \includegraphics[width=\linewidth]{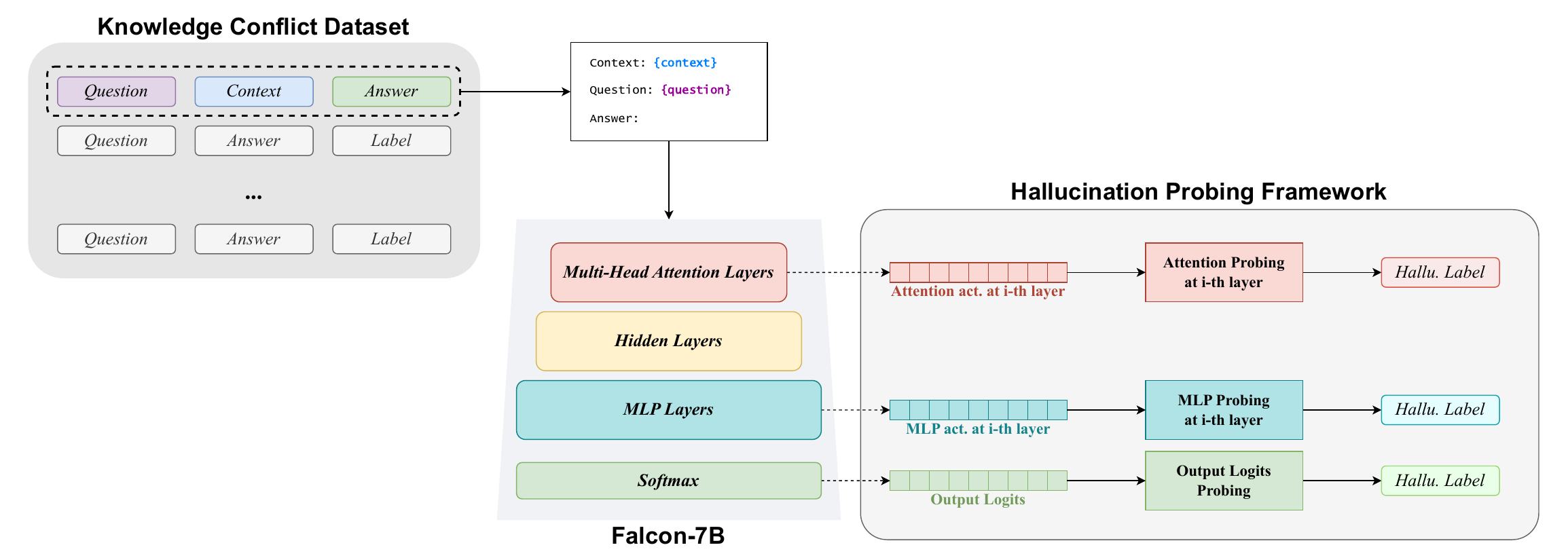}
    \caption{Probing knowledge conflicts through hallucinations. The NQ-Swap dataset is processed using Falcon-7B, from which output logits and intermediate activations (attention and MLP layers) are extracted. These are passed to dedicated probing classifiers to predict hallucination labels, which are compared with the ground-truth knowledge conflict labels.}
    \label{fig:kc_by_hall}
\end{figure}

The whole framework is shown in Fig.~\ref{fig:kc_by_hall}.

\section{Results}
\label{sec:results}

In this section, we empirically examine the relationship between hallucinations and knowledge conflicts in LLMs. We begin by evaluating whether knowledge conflict signals can be effectively used to detect hallucinations. We then explore the inverse relation—whether hallucination signals can serve as reliable indicators of knowledge conflicts. Finally, we assess the robustness of these probing models across different languages to test their generalizability. 

\subsection{Probing Hallucinations through Knowledge Conflicts}
\label{subsec:hall_by_kc}

To assess whether hallucinations stem from knowledge conflicts, we implemented the pipeline represented in Fig.~\ref{fig:hall_by_kc}. We analyzed whether hallucination signals are encoded similarly to those of knowledge conflicts within LLaMA-3-8B, so that the former can be detected by using the latter.

Following the proposal in \cite{zhao2024analysingresidualstreamlanguage}, the knowledge conflict probing models were implemented as Logistic Regression classifiers. All models were trained with the Adam optimizer (learning rate \(0.002\)) and a StepLR scheduler with a decay factor of 0.95 per epoch, using a batch size of 64 and 20 training epochs. An \(L_1\) regularization term (\(0.0002 \times 3\)) was also added. Training was repeated 20 times for robustness, and performance was assessed using accuracy and AUROC.

\begin{figure}[t]
    \centering
    \begin{subfigure}[b]{0.4\textwidth}
        \includegraphics[width=\textwidth]{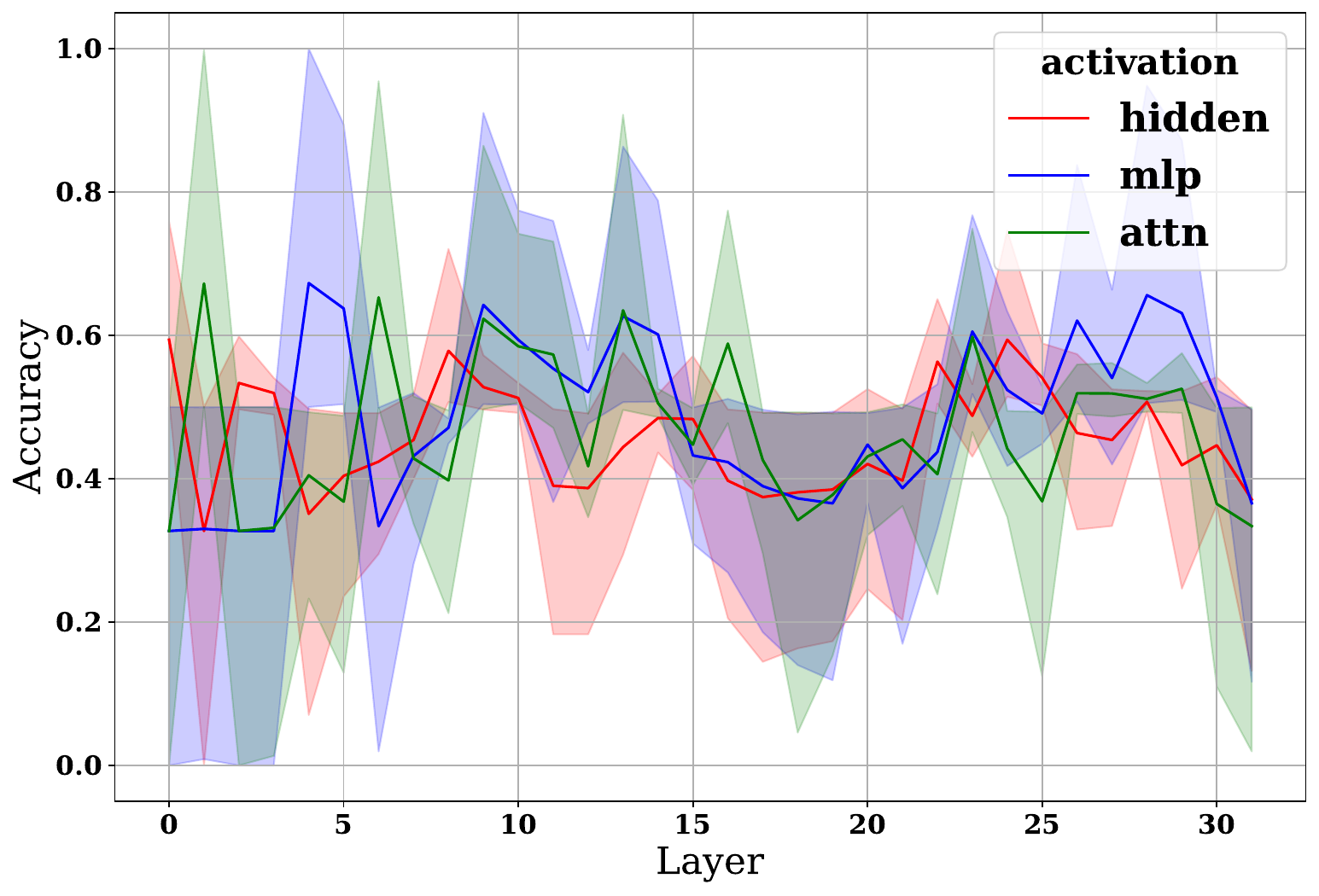} 
        \caption{Accuracy}
    \end{subfigure}
    \begin{subfigure}[b]{0.4\textwidth}
        \includegraphics[width=\textwidth]{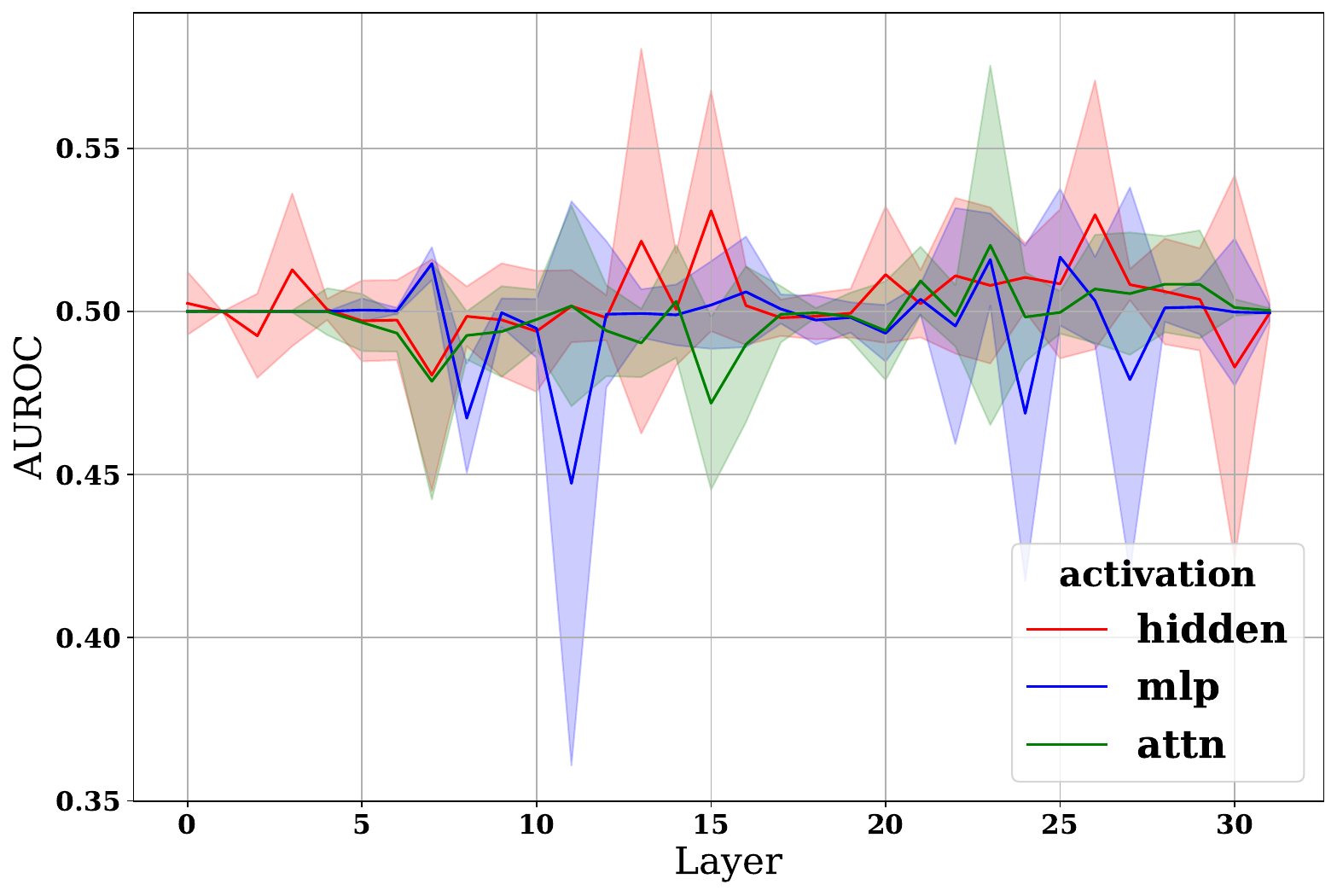}
        \caption{AUROC}
    \end{subfigure}
    \caption{
        Accuracy and AUROC of knowledge conflict probing models on detecting hallucinations based on the activations of LLaMA-3-8B. The probing results for hidden, attention, and MLP activations are colored red, green, and blue, respectively.}
    \label{fig:hallucination_kc_results}
\end{figure}

Figure \ref{fig:hallucination_kc_results} shows the layer-wise performance across three activation types: hidden, attention, and MLP. We observe that accuracy values fluctuate across layers but remain above \(0.4\) in several layers, generally for all activation types. Notably, a decline in accuracy in detecting hallucinations was observed, particularly where knowledge conflicts were detected more precisely (from layer 15 to 25). Furthermore, we noticed relatively higher performances in the initial and final layers, achieving \(\sim0.65\) of accuracy at layers \(4\) and 28 with MLP activations, and at layers 1 and 6 with attention activations. Another important difference with respect to the reference study \cite{zhao2024analysingresidualstreamlanguage} is that hidden activations seem to be less informative than the others.
AUROC trends are consistent with the accuracy findings, though the overall AUROC values remain close to 0.5, indicating limited discriminative ability.

These findings indicate that the internal representations encoding knowledge conflicts in LLaMA-3-8B do not contain meaningful signals for hallucination detection. In other words, within our experimental setting, probing models trained on knowledge conflict-related activations appear largely ineffective at predicting hallucinations. This suggests that, at least in this context, the two phenomena are not strongly aligned, providing empirical evidence that hallucination representations may not be fully reducible to, or explained by, knowledge conflict representations.

\subsection{Probing Knowledge Conflicts through Hallucinations}
\label{subsec:kc_by_hall}

To test the inverse relationship  (i.e., whether knowledge conflicts are captured as hallucinations are), we implemented the pipeline represented in Fig. \ref{fig:kc_by_hall}. So, we analyzed whether knowledge conflict signals are encoded similarly to those of hallucinations within Falcon-7B, so that the former can be detected by using the latter.

Following what was proposed in \cite{snyder2024early}, the hallucination probing models were implemented using feed-forward classifiers consisting of a linear layer with 256 units, followed by a ReLU activation, a dropout layer with a rate of 0.5, and a final classification layer. All models were trained with the AdamW optimizer (learning rate \(1 \times 10^{-4}\), weight decay \(1 \times 10^{-2}\)), a batch size of 128, and 1000 training steps. Data was split into training and test sets with an 80-20 ratio, and performance was assessed using AUROC and accuracy.

\begin{figure}[ht]
    \centering
    \begin{subfigure}[b]{0.4\textwidth}
        \includegraphics[width=\textwidth]{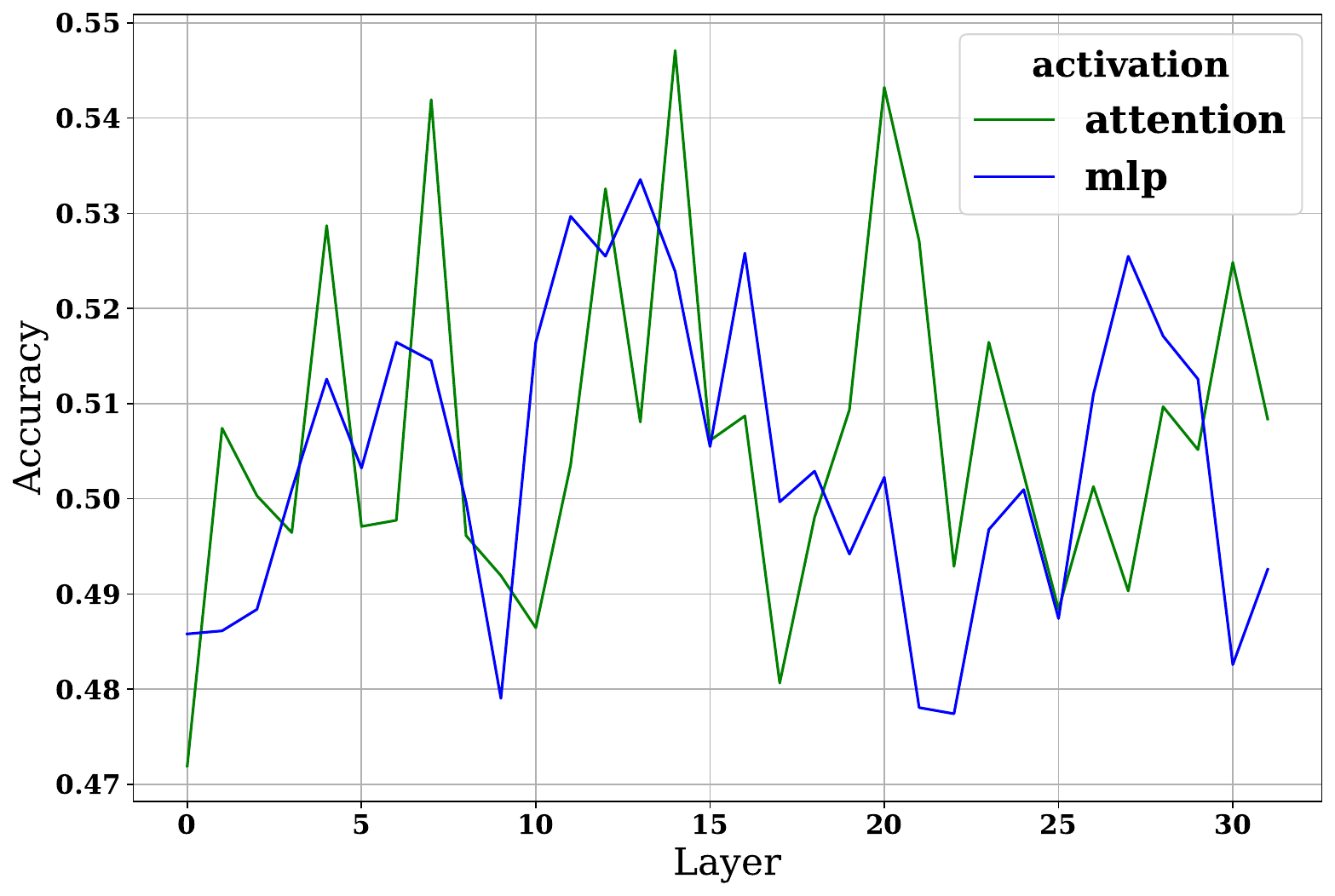} 
        \caption{Accuracy}
    \end{subfigure}
    \begin{subfigure}[b]{0.4\textwidth}
        \includegraphics[width=\textwidth]{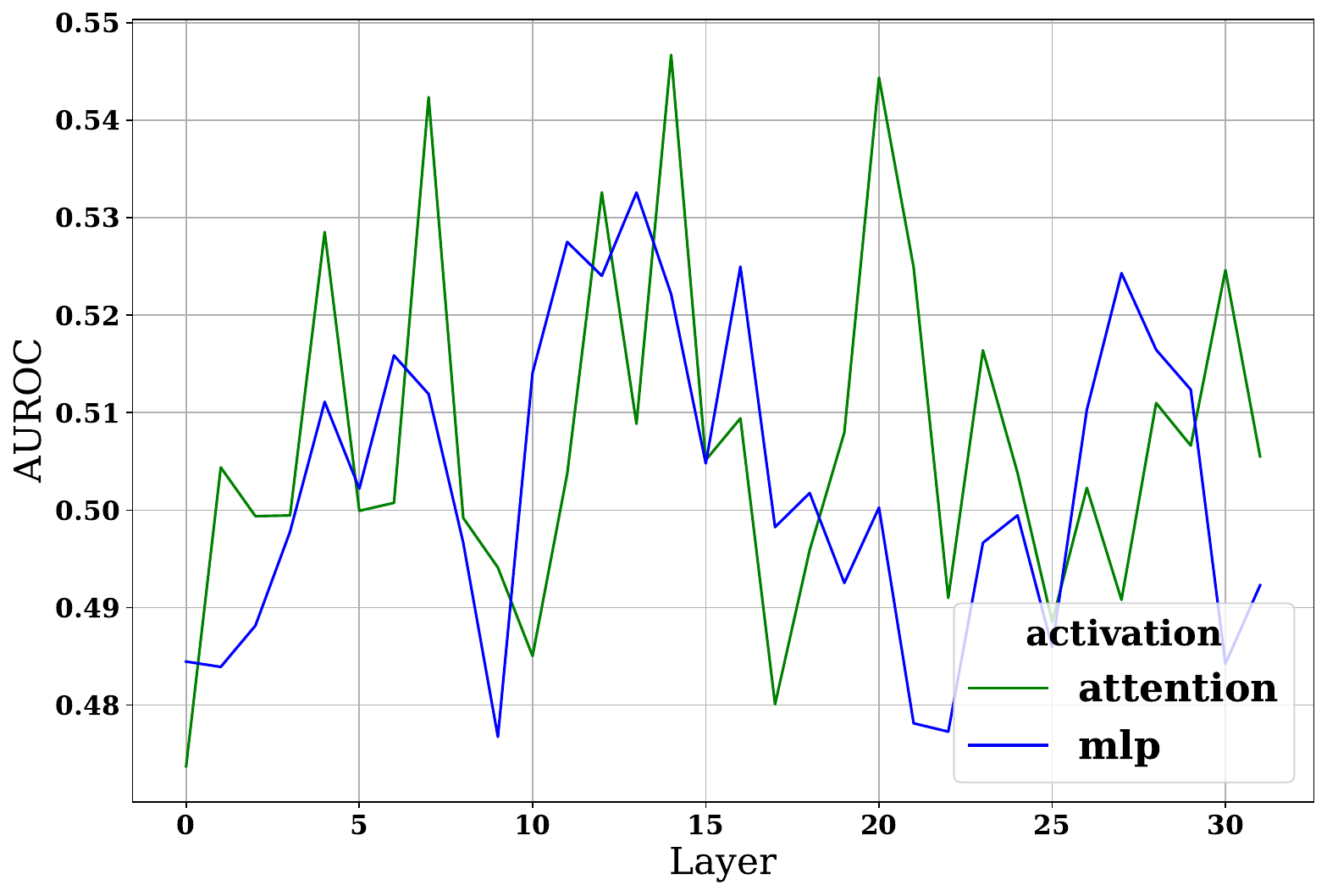}
        \caption{AUROC}
    \end{subfigure}
    \caption{
        Accuracy and AUROC of hallucination probing models on detecting knowledge conflicts based on the activations of Falcon-7B. The probing results for attention and MLP activations are colored green and blue, respectively.}
    \label{fig:kc_hallucination_results}
\end{figure}

As shown in Fig.~\ref{fig:kc_hallucination_results}, both metrics remain close to chance level throughout the network. Accuracy hovers around 0.5 with minor fluctuations, and AUROC curves closely follow a similar trend, indicating limited discriminative ability. While there are occasional spikes, particularly in intermediate layers 14 and 20, these reach only \(\sim 0.55\), thus resulting negligible. This behavior significantly differs from that observed in the reference research \cite{snyder2024early}, where hallucination detection improves substantially towards the final layers. Another difference is that, in \cite{snyder2024early}, MLP activations are generally more informative than the attention ones, while here, the opposite happens. This also holds when probing output logits, where we registered a decay of \(10\%\) in performance, as observed from Table \ref{tab:probing_logits_performance}.

These findings indicate that the internal representations encoding hallucinations in Falcon-7B do not contain meaningful signals for knowledge conflict detection. In our experimental setting, this suggests that the two phenomena are not strongly aligned, providing empirical evidence that hallucination representations may not directly account for knowledge conflict representations.  
 
\begin{table}[ht]
    \caption{Accuracy and AUROC of the probing models proposed in \cite{snyder2024early} in detecting hallucinations and knowledge conflicts by using output logits.}
    \label{tab:probing_logits_performance}
    \begin{tabular}{lccc}
        \toprule
        \textbf{Task} & \textbf{Accuracy} & \textbf{AUROC} & \textbf{Dataset} \\
        \midrule
        Hallucination detection & 0.626 & 0.655 & TriviaQA \\
        Knowledge conflict detection & 0.519 & 0.517 & NQ-Swap \\
        \bottomrule
    \end{tabular}
\end{table}

\subsection{Probing Method Robustness}
\label{subsec:multi_lang}

\begin{figure}[t]
    \centering
    \begin{subfigure}[b]{0.33\textwidth}
        \includegraphics[width=\textwidth]{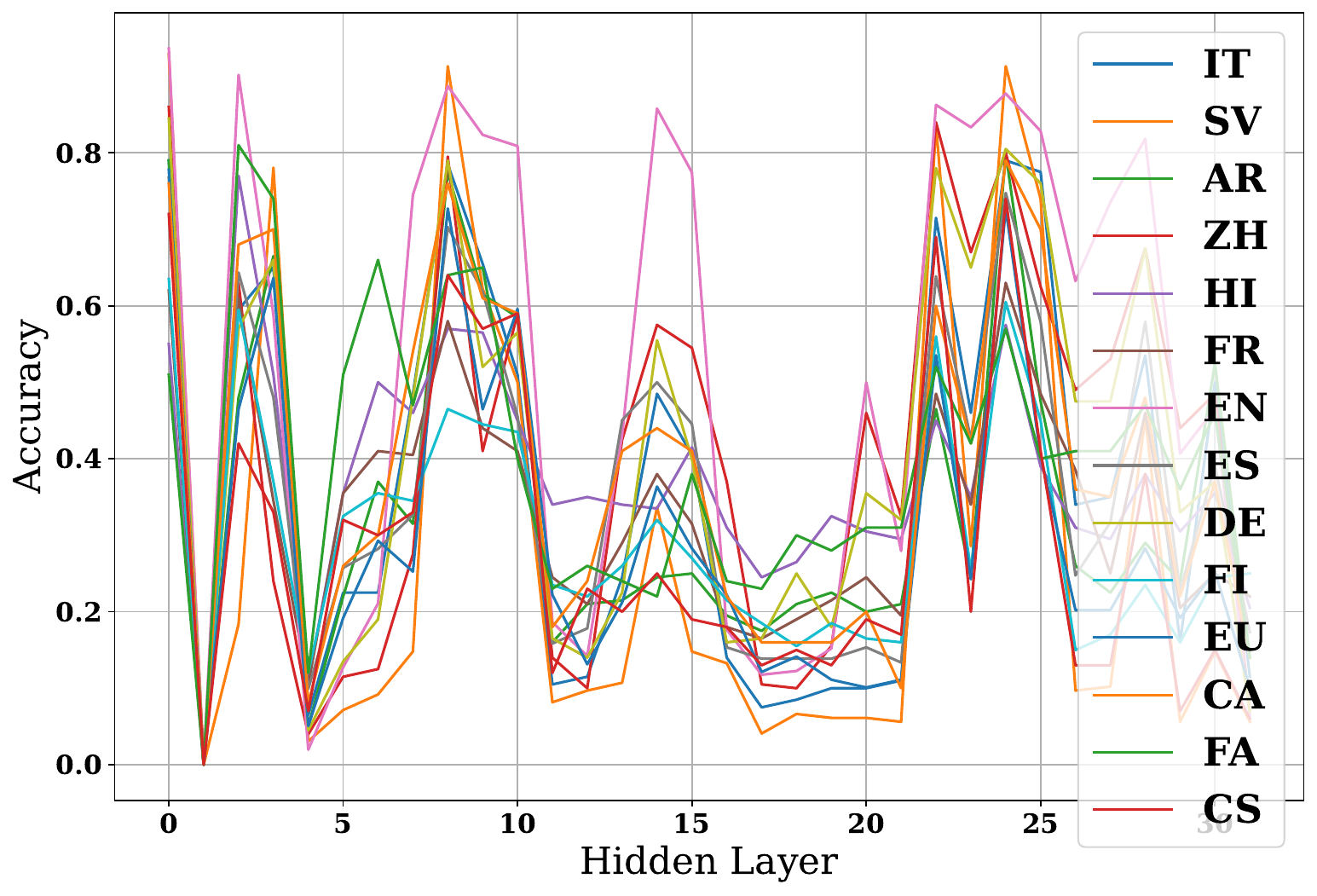} 
        \caption{Hidden activations}
    \end{subfigure}
    \begin{subfigure}[b]{0.33\textwidth}
        \includegraphics[width=\textwidth]{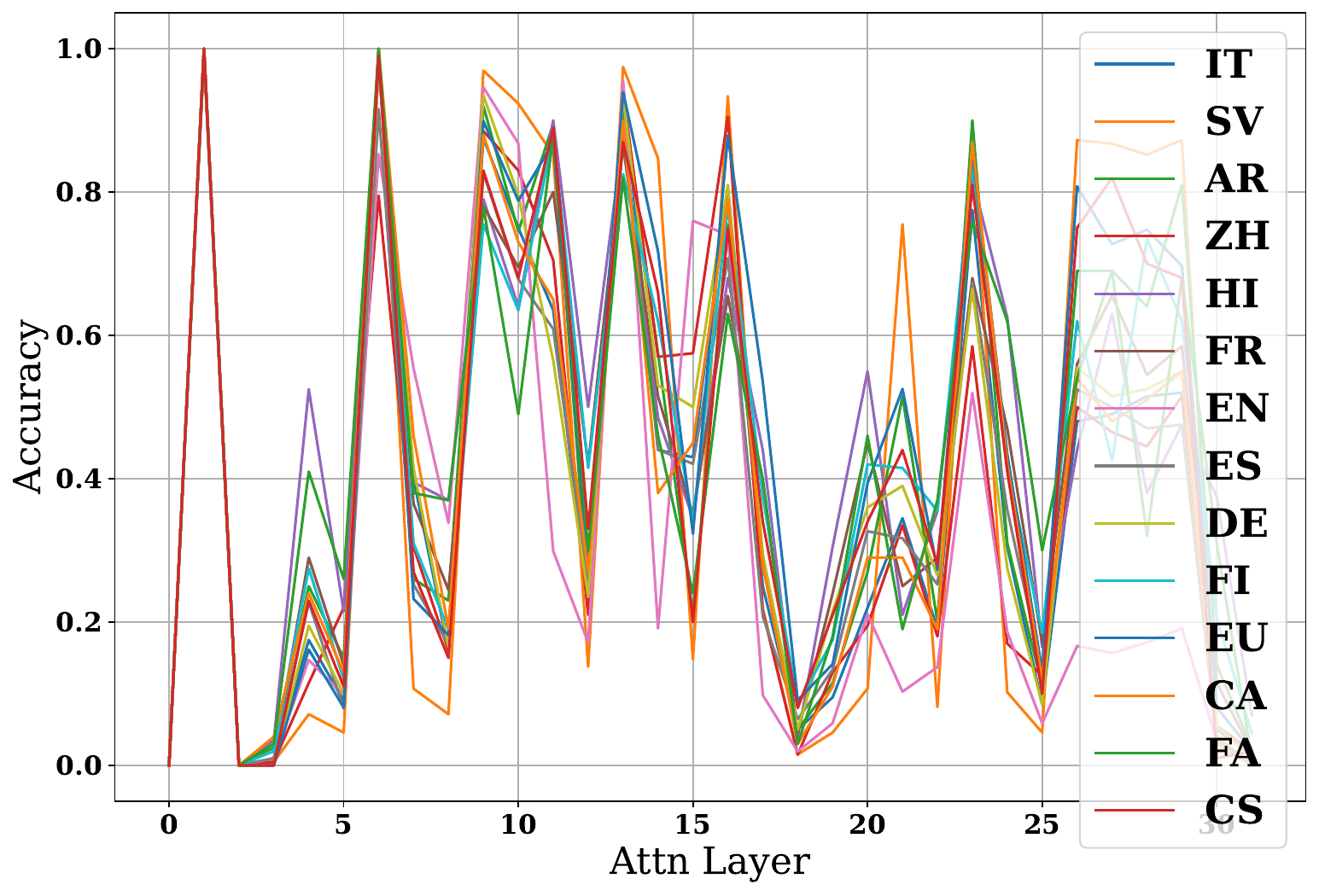}
        \caption{Attention activations}
    \end{subfigure}
    \begin{subfigure}[b]{0.33\textwidth}
        \includegraphics[width=\textwidth]{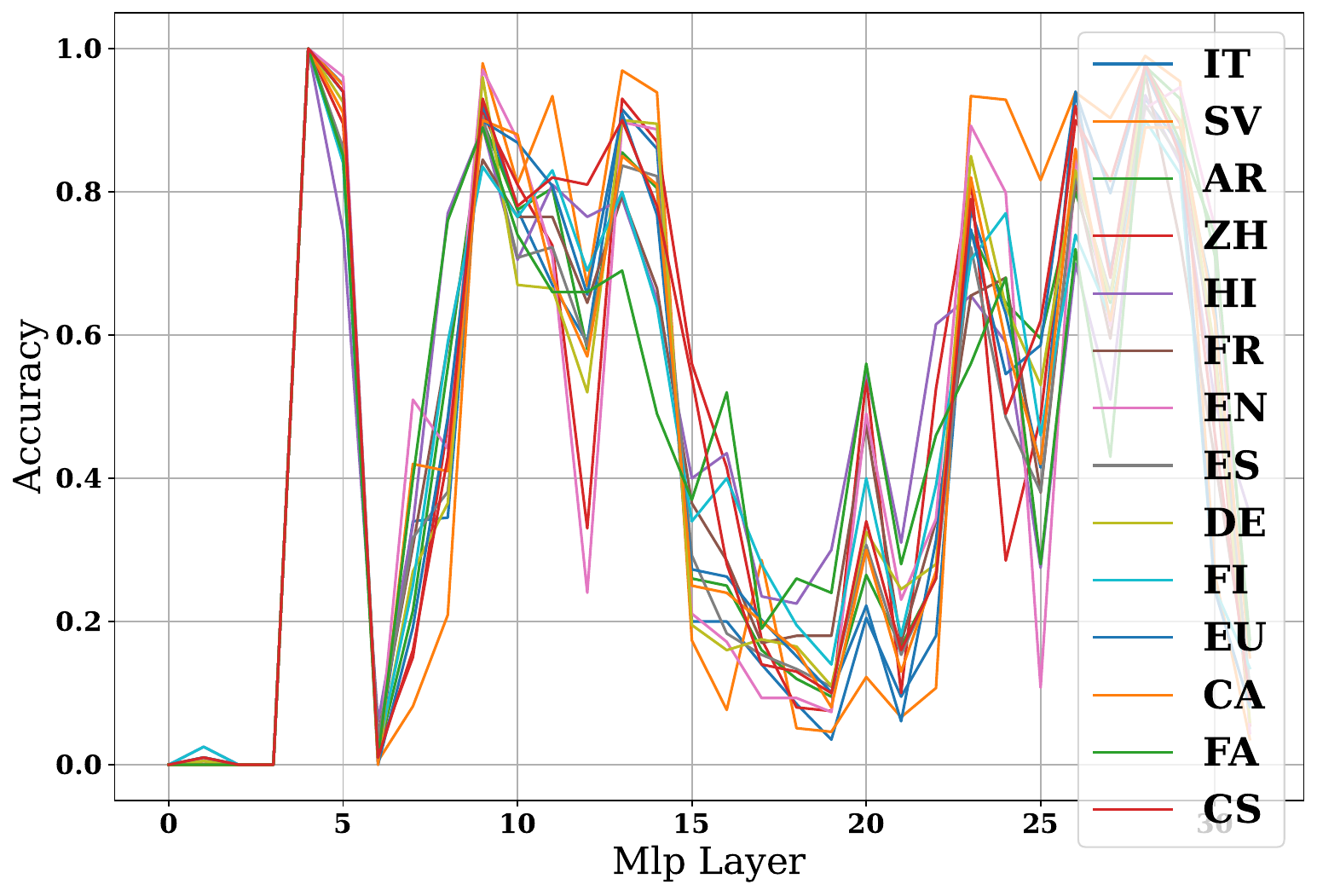}
        \caption{MLP activations}
    \end{subfigure}
    \caption{
        Accuracy of knowledge conflict probing models in detecting hallucinations across 14 languages, based on LLaMA-3-8B activations. Probing results are derived from the Mu-SHROOM dataset. }
    \label{fig:multi_lang_results}
\end{figure}

To evaluate the robustness of the probing method from a general point of view, we analyze the performances of knowledge conflict probing models applied to the Mu-SHROOM dataset (since it was the only dataset including non-English instances), as illustrated in Fig.~\ref{fig:hall_by_kc}, across different activation types, model layers, and languages. For each examined activation of LLaMA-3-8B, the accuracy across \(14\) languages is shown in Fig.~\ref{fig:multi_lang_results}.

A key observation is that probing accuracy remains generally stable across languages, particularly for attention and MLP activations. Despite minor fluctuations, most language curves cluster closely, indicating that the probing method is not heavily biased toward any specific language, supporting its multilingual robustness.

Regardless of our specific probing setup and the performance levels observed, each reference work \cite{zhao2024analysingresidualstreamlanguage, snyder2024early} achieved encouraging results. This highlights the effectiveness of the underlying probing technique, especially when considering its consistent behavior across both high-resource and low-resource languages.

\section{Conclusion and Future Works}
\label{sec:conclusion}

In this paper, we presented a novel investigation into the relationship between hallucinations and knowledge conflicts—two phenomena that, while often considered intertwined, lack systematic empirical analysis. Motivated by the hypothesis that internal knowledge inconsistencies may underlie model hallucinations, we explored the bidirectional relationship between them through a probing-based framework. To evaluate whether hallucinations stem from knowledge conflict, we applied the knowledge conflict probing approach proposed by Zhao et al.~\cite{zhao2024analysingresidualstreamlanguage} to three benchmark datasets for hallucination detection, using LLaMA-3-8B and extracting hidden, attention, and MLP activations across all layers. Conversely, to test whether knowledge conflicts are captured like hallucinations, we employed hallucination probing techniques following Snyder et al.~\cite{snyder2024early}, analyzing output logits and intermediate representations from the attention and MLP layers of Falcon-7B on a benchmark dataset for knowledge conflict. 

Our empirical findings indicate that, within our experimental setting, no significant correlation was observed between hallucinations and knowledge conflicts at the level of internal representations, despite the intuitive assumption of a strong causal link. This suggests that hallucinations may arise from more complex or orthogonal mechanisms, and that knowledge conflict alone is not a sufficient predictor. Nonetheless, our analysis reveals that probing models remain consistent across linguistic variations, maintaining stable performance even in non-English settings. This supports their potential utility in multilingual and globally distributed AI systems. While the lack of correlation may seem discouraging, it contributes valuable clarity to ongoing discussions about the interpretability and diagnostic capabilities of LLM internals. It also highlights the need to decouple these phenomena in future research, thereby encouraging more targeted approaches for hallucination mitigation.

Looking ahead, we propose extending this probing-based methodology to domain-sensitive applications such as digital health \cite{VALERIO2025108922, ullah2024challenges, sarker2024natural}, education \cite{ho2024mitigating, neumann2024llm}, and human resource management \cite{chkirbene2024large, laraspata2024enhancing}, where the reliability of LLM outputs is critical. These domains present unique challenges not only in controlling hallucinations but also in ensuring transparency, fairness, and user trust. Moreover, ethical considerations are crucial: misinterpreting probing results may lead to flawed interventions, biases in multilingual datasets can disadvantage underrepresented groups, and misclassifying hallucinations in high-stakes contexts—such as medical or hiring decisions—could have serious consequences. Addressing these risks requires not only technical rigor but also an integrated approach that accounts for ethical, social, and regulatory dimensions. 

Finally, our experiments were conducted using a single LLM per task. To assess the generalizability of our findings, future work should involve broader comparisons across multiple state-of-the-art language models, potentially uncovering architecture-specific behaviors and further informing the design of trustworthy LLM applications.

\begin{acknowledgments}
A Ph.D. fellowship funds Lucrezia Laraspata’s research within the Italian “D.M. n. 630, April 24, 2024” – under the NRRP, Mission 4, Component 2, Investment 3.3 – Ph.D. project “Development and application of Artificial Intelligence methods for Human Capital Management”, co-supported by Talentia Software s.r.l. (CUP B91I24000240007).

Giovanna Castellano acknowledges funding from the FAIR - Future AI Research project, Spoke 6 - Symbiotic AI (CUP H97G22000210007), under the NRRP MUR program funded by NextGenerationEU. 

We acknowledge ISCRA for awarding this project access to the LEONARDO supercomputer, owned by the EuroHPC Joint Undertaking, hosted by CINECA (Italy).

\end{acknowledgments}

\section*{Declaration on Generative AI}

We used GPT-4o and Grammarly to assist with grammar refinement, spelling correction, and stylistic improvements during the preparation of this manuscript. All content was subsequently reviewed, revised, and approved by the authors, who take full responsibility for the final version of the work.

\bibliography{sample-ceur}

\end{document}